\documentclass{article}
\usepackage{spconf,amsmath,graphicx}
\usepackage{array,multirow,makecell}
\usepackage{multicol,multirow}
\usepackage[center]{caption}
\usepackage{subcaption}
\usepackage{hyperref}

\title{Benchmarking performance of object detection under image distortions in an uncontrolled environment}
\name{Ayman Beghdadi, Malik Mallem and Lotfi Beji}
\address{IBISC Lab, Univ Evry, Paris-Saclay University, Evry, France\\
	aymanaymar.beghdadi@univ-evry.fr, malik.mallem@univ-evry.fr, lotfi.beji@univ-evry.fr}

\begin{document}
\topmargin=0mm 
\ninept
\maketitle
\begin{abstract}
The robustness of object detection algorithms plays a prominent role in real-world applications, especially in  uncontrolled environments due to distortions during image acquisition. It has been proven that the performance of object detection methods suffers from in-capture distortions. In this study, we present a performance
evaluation framework for the state-of-the-art object detection methods using a dedicated dataset containing images with various distortions at different levels of severity. Furthermore, we propose an original strategy of image distortion generation applied to the MS-COCO dataset that combines some local and global distortions to reach much better performances. We have shown that training using the proposed dataset improves the robustness of object detection  by 31.5\%. Finally, we provide a custom dataset including  natural images distorted from MS-COCO to perform a more reliable evaluation of the robustness against common distortions. The database and the generation source codes of the different distortions are made publicly available\footnote{\href {https://github.com/Aymanbegh/Benchmarking-performance}{https://github.com/Aymanbegh/Benchmarking-performance}}$^{,}$\footnote{\href {https://github.com/Aymanbegh/Distorted-Natural-COCO}{https://github.com/Aymanbegh/Distorted-Natural-COCO}}.
\end{abstract}

\section{Introduction}
\label{sec:intro}

Object detection (OD) and recognition is one  of the most challenging and widely studied problems in computer vision and especially in uncontrolled environment.
Deep Neural Networks (DNN) based approaches \cite{zou2019object,zhao2019object,liu2020deep} have been proven very promising paradigms to solve such complex problems in various real-world applications. However, generally, the proposed DNN-based architectures do not consider the sensitivity of the learning models to common data distortions and corruption effects. Indeed, it has been shown by Szegedy et al.\cite{szegedy2013intriguing} that a small perturbation invisible to the human visual system can have significant repercussions on the performance of object recognition models. It is therefore important to conduct ad hoc studies to understand the effects of such perturbations 
and to be able to generate them at different levels in order to develop more robust OD networks.
The limitations of object detection methods, in terms of robustness against some common signal degradation, data corruptions, or adversarial examples, have been highlighted in several works \cite{michaelis2019benchmarking, hendrycks2019benchmarking,li2021universal,benz2021revisiting}.
A solution would be to perform pre-processing before the high-level phases. But this implies being able to identify the distortion automatically in order to apply the most appropriate solution. A more elegant and efficient solution present in many research \cite{khodabandeh2019robust,geirhos2018imagenet,cygert2021robustness,poursaeed2021robustness} is to enrich the database used for learning by simulating the most relevant distortions. It is in this last framework that the proposal described in this work is made.
Thereby, we propose a full framework evaluation of robustness for a set of object detection methods \cite{bochkovskiy2020yolov4,tan2020efficientdet,he2017mask} through several distortions applied on the MS-COCO dataset \cite{lin2014microsoft}. 
This data augmentation is performed through some common global distortions (noise, motion blur, defocus blur, haze, rain, contrast change, and compression artefacts) in the whole image and some local distortions (local blur motion, BackLight illumination and local defocus blur)  in specific areas that include the possible dynamic objects or scene conditions. 
To the best of our knowledge, our work is the first to consider the local distortions to assess the robustness of OD models.
Finally, we proceeded to a manual selection of the distorted images present in the validation set of the MS-COCO dataset and retained only the annotations of the distorted objects to perform a more relevant evaluation of the impact of distortions on models.
The main contributions of our study are summarized as follows:
\begin{itemize}
\item A comprehensive evaluation of the robustness of the state-of-the-art OD methods against global and local distortions at 10 levels of distortion is conducted (see section \ref{Exp}.1).
\item A dataset dedicated to the study of the impact of local and global distortions on the robustness of OD is built from MS-COCO dataset \cite{Begh2022Bench}.
\item It is shown that the robustness of OD process is improved by enriching the training process  using data augmentation based on generated distorted images (see table \ref{Training}).
\item An additional new benchmark dataset constructed for natural in-capture distortions from MS-COCO dataset that allows to better assess the reliability of the OD models  in case of real distortions is provided (see tables  \ref{Tools} and \ref{Training2}) \cite{Begh2022DN-COCO}. 
\end{itemize}
The remainder of the paper is organized as follows. Section \ref{Related} summarizes previous related literature. Section
\ref{methods} is devoted to detail the methods of evaluation and the distortion generations. Then, section \ref{Exp} is dedicated to show and discuss results.
Finally, concluding remarks and perspectives are provided in section \ref{conclusion}.
\begin{figure*}[t]
    \centering
    \begin{subfigure}[b]{0.24\textwidth}
        \centering
        \includegraphics[width=\textwidth]{normal.jpg}
        \caption{Original}
    \end{subfigure}
    \hfill
    \begin{subfigure}[b]{0.24\textwidth}
        \centering
        \includegraphics[width=\textwidth]{defocus.jpg}
        \caption{Local defocus blur}
    \end{subfigure}
    \hfill
    \begin{subfigure}[b]{0.24\textwidth}
        \includegraphics[width=\textwidth]{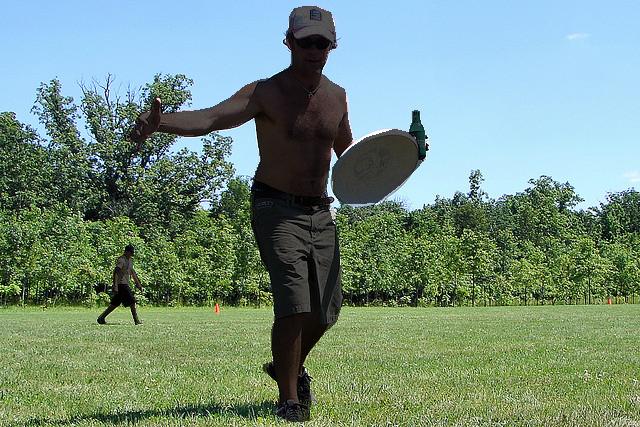}
        \caption{BackLight illumination (BL)}
    \end{subfigure}
    \hfill
    \begin{subfigure}[b]{0.24\textwidth}
        \centering
        \includegraphics[width=\textwidth]{blur.jpg}
        \caption{Local motion blur}
    \end{subfigure}
    \caption{Illustration of some local distortions.}
    \label{distortion}
\end{figure*}

\section{Related Work}
\label{Related}
The evaluation of OD methods from images acquired in an uncontrolled environment has been the subject of a few studies \cite{cygert2020toward,chen2021robust}. But most of the time, the used databases contain images with limited number of degradation types and therefore cannot be generalized to practical cases where one has to face several distortions at the same time.
Some papers deal with the impact of various distortions on the detection performance, described in \cite{azulay2018deep}, that attempts to evaluate the robustness of some backbones of CNN architectures against geometric transformations. A comparative study between DNN-based architectures
(ResNet-152, VGG-19, GoogLeNet) 
and human observer performance was conducted in \cite{geirhos2018generalisation}. In this study  twelve common distortions are considered in the performance evaluation of object recognition methods. It has been shown that these DNN-based architectures perform remarkably well in the case of the distortions on which the training was performed. However, this performance decreases, compared to that of the human observer, for unseen distortions.
Another interesting study on the evaluation of the robustness of classification models of images from the imagenet database and subjected to fifteen artificial distortions at five levels of severity, revealed the effect of image quality degradation on the stability of the training process for various DNN-based architectures \cite{hendrycks2019benchmarking}.
Recently, a benchmark study on the robustness of many OD models against 15 types of global distortions and stylized images through data augmentation process has been done in \cite{michaelis2019benchmarking}.
\newline
\section{Proposed benchmarking framework}
\label{methods}
In order to evaluate the robustness of the considered models against image distortions in an uncontrolled environment, we propose 10 types of distortions (global and local) at 10 linearly increasing levels of severity \cite{Begh2022Bench}. 
In figures \ref{distortion} (b), (c), and (d), distortion levels are 8, 7, and 10 respectively. 

\subsection{Image distortions}
The distortions of the image signal can be classified into two main categories described below.
\textbf{Global distortions} affect the image as a whole and come from different sources related in general to the acquisition conditions. Some are directly dependent on the physical characteristics of the camera and are of photometric or geometric origin. Among the most common distortions that affect the quality of the signal are defocus blur (Defocus-Blur), photon noise, geometric or chromatic aberrations, and blur (Motion-Blur) due to the movement of the camera. The other types of degradation are related to the environment and, more particularly, the lighting and atmospheric disturbances in the case of outdoor scenes. Compression and image transmission artifacts are another source of degradation that is difficult to control. These common distortions have been already considered in benchmarking the performance of some models \cite{michaelis2019benchmarking,hendrycks2019benchmarking,benz2021revisiting}.
\textbf{Local distortions} are undesirable signals affecting one or more localized areas in the image (see figure \ref{distortion}). A typical case is the Motion Blur (Loc. MBlur) due to the movement of an object of relatively high speed. 
Another photometric distortion is the appearance of a halo around the object contours due to the limited sensitivity of the sensors or backlight illumination (BackLight). The artistic blur (Loc. Defoc.) affecting a particular part of the targeted scene, the object to be highlighted by the pro-shooter, is another type of local distortion. 
Thus, integrating the local distortions in the database increases its size and makes it richer and more representative of scenarios close to real applications.
It is worth noticing that these local distortions are made possible thanks to the annotations provided in the MS-COCO database and, in particular, the details of the shape and location of the object of interest.

\subsection{Description of the considered models }

 In this study we focus on few models of state-of-the-art object detection methods which have the best performance or notoriety, namely YOLOv4 \cite{bochkovskiy2020yolov4}, Mask R-CNN \cite{he2017mask} 
\begin{table}[h]
\caption{State-of-the-art object detection models.}
\begin{center}
\begin{tabular}{p{2.4cm}p{2.1cm}p{0.4cm}p{1.9cm}}
\hline\noalign{\smallskip}
Methods&Architecture&Size&Add-blocks\\
\noalign{\smallskip}\hline
\noalign{\smallskip}\hline\noalign{\smallskip}
\multirow{2}{*}{Mask R-CNN \cite{he2017mask}}&\multirow{2}{*}{Resnet-101}&\multirow{2}{*}{512}&FPN,  RPN \cite{lin2017feature,ren2015faster}\\
\hline\noalign{\smallskip}
\multirow{2}{*}{YOLOv4 \cite{bochkovskiy2020yolov4}}&CSPDarknet53&512&\multirow{2}{*}{PAN \cite{liu2018path}}\\
&YOLOv4-tiny&416&\\
\hline\noalign{\smallskip}
\multirow{5}{*}{EfficientDet \cite{tan2020efficientdet}}&EfficientNet-B0&512&\multirow{5}{*}{Bi-FPN \cite{tan2020efficientdet}}\\
&EfficientNet-B1&640&\\
&EfficientNet-B2&768&\\
&EfficientNet-B3&896&\\
&EfficientNet-B4&1024&\\
\noalign{\smallskip}\hline\noalign{\smallskip}
\end{tabular}
\end{center}
 FPN: Feature Pyramid Network, RPN: Region Proposal Network, PAN: Path Aggregation Network, Bi-FPN: Bi-directional FPN.
\label{Obj_detect}
\end{table}
and EfficientDet \cite{tan2020efficientdet} with distinct backbone architectures that highlights the correlation between the backbone and their robustness. The table \ref{Obj_detect} summarizes the different models and their components which are used in this benchmarking.

\subsection{Robustness Evaluation against Natural Distortions}
\begin{figure}[t]
    \centering
    \begin{subfigure}[b]{0.48\linewidth}
        \centering
        \hspace*{-0.6cm}\includegraphics[width=1.3\linewidth]{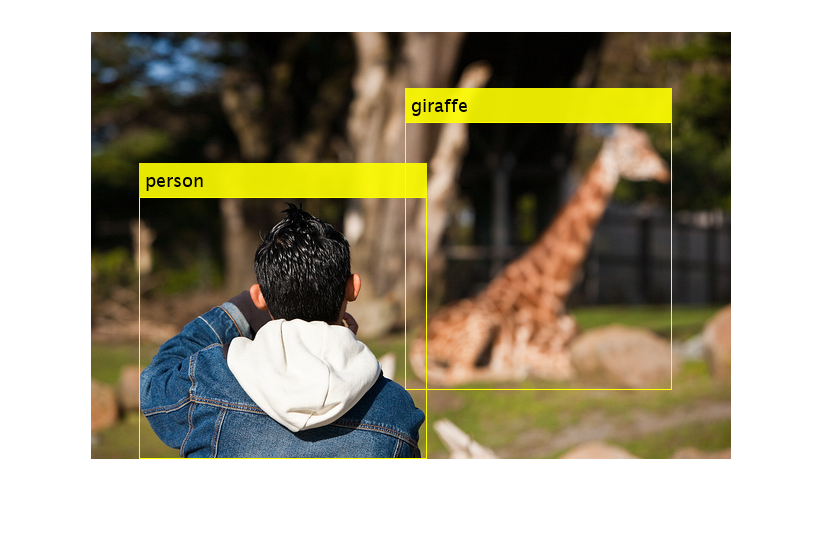}
        \caption{Original annotations}
    \end{subfigure}
    \hfill
    \begin{subfigure}[b]{0.48\linewidth}
        \centering
        \hspace*{-0.6cm}\includegraphics[width=1.3\linewidth]{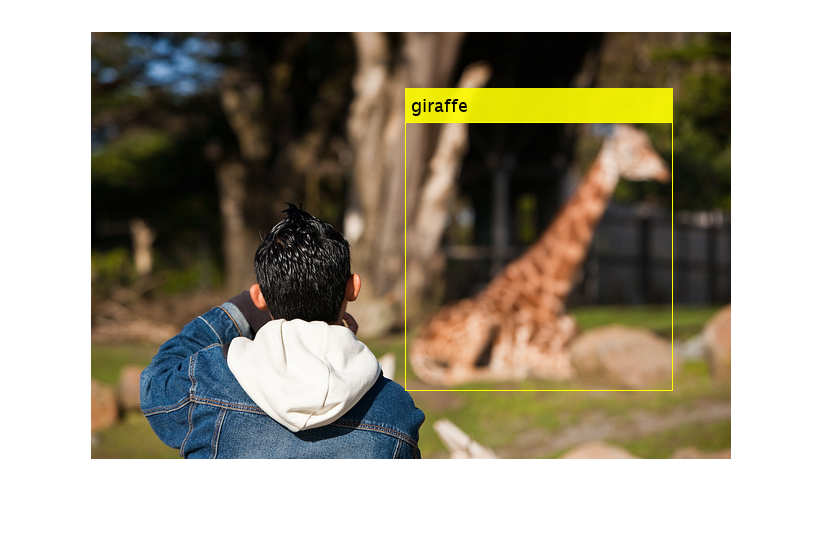}
        \caption{Sorted annotations}
    \end{subfigure}
    \caption{Natural distortions selection in the MS-COCO dataset.}
    \label{selection}
\end{figure}

We manually identified the natural distortions present in the COCO validation set to better assess the robustness in real scenarios. Indeed, in the existing methods, robustness evaluations were performed on synthetic distortions or specific datasets, thus limiting those studies' extension in real-world applications.
It is worth noticing that the selected sub-sets from COCO contain images with global distortions and the associated object annotations. In contrast, only affected objects are considered in the case of local distortions (see figure \ref{selection}).
This benchmarking framework for robustness evaluation uses one of the most widespread databases for OD, restricted to a limited set of seven natural distortions types (see table \ref{Tools}).
In this way, we aim to lay the foundations for a future common evaluation method  for evaluating OD robustness against real-world distortions.
\begin{table}[h]
\caption{Features of distorted natural sub-sets.}
\begin{center}
\begin{tabular}{p{0.9cm}|p{0.4cm}p{0.8cm}p{0.3cm}p{0.6cm}p{0.5cm}p{0.6cm}p{0.9cm}}
\hline
& Noise& Contrast& Blur& Defoc.& Rain& O.Blur& BackL.\\

\hline\noalign{\smallskip}
\hline
Images&44&42&32&201&21&127&128\\
Objects&289&312&224&1299&110&464&934\\
Ratio&1.0&0.83 &0.97&0.72&0.95&0.34&0.68\\
\hline
\end{tabular}
\end{center}
\label{Tools}
Ratio: Number of retained objects/annotated objects in images.
\end{table}
It is worth  noting that some natural distortions such as Defocus, Object Blur, and BackLight have low ratios (see table  \ref{Tools}). This results from the nature of local distortions, which only affect specific objects in the images (see figure \ref{selection}).

\section{Results and discussions}
\label{Exp}
\subsection{Evaluation of the robustness}
\begin{figure}[h]
    \centering
        \centering
        \includegraphics[width=0.9\linewidth]{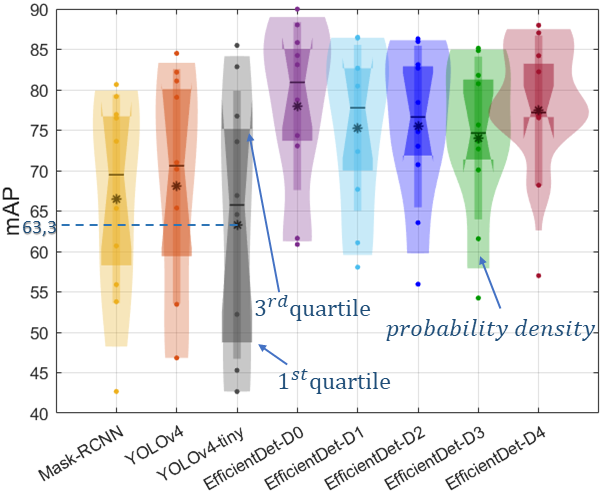}
        
        */-: Mean/Median of mAP for all distortion type and level, Light/dark surrounding shape: probability density/percentiles. 
    \caption{Average robustness rate of object detection models against the distortions.}
    \label{rate}
\end{figure}

We evaluated the robustness of models against the distortions on the MS-COCO 2017 validation set, 
\label{Exp_eval}
\begin{figure}[h]
    \centering
        \centering
        \includegraphics[width=\linewidth]{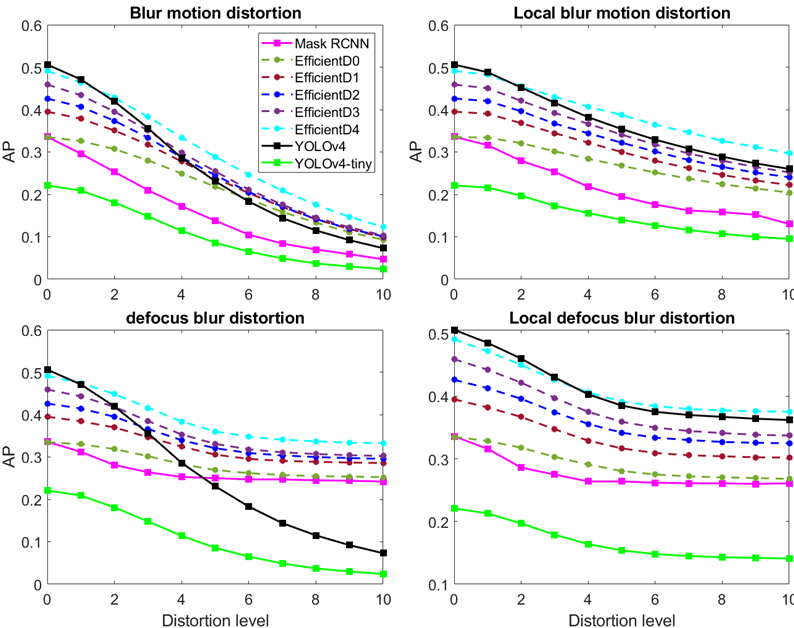}
    \caption{Robustness of object detection models against global(left)/local (right) distortions.}
    \label{eval2}
\end{figure}
which contains 5K annotated images and corrupted by the 10 types of distortions we generated \cite{Begh2022Bench}. it is worth noting that level 0 of distortion corresponds to  non-distorted images.
We assessed the  performance of the ODn algorithms using the AP (Average Precision) metric.
\begin{figure}[h]
    \centering
        \centering
        \includegraphics[width=\linewidth]{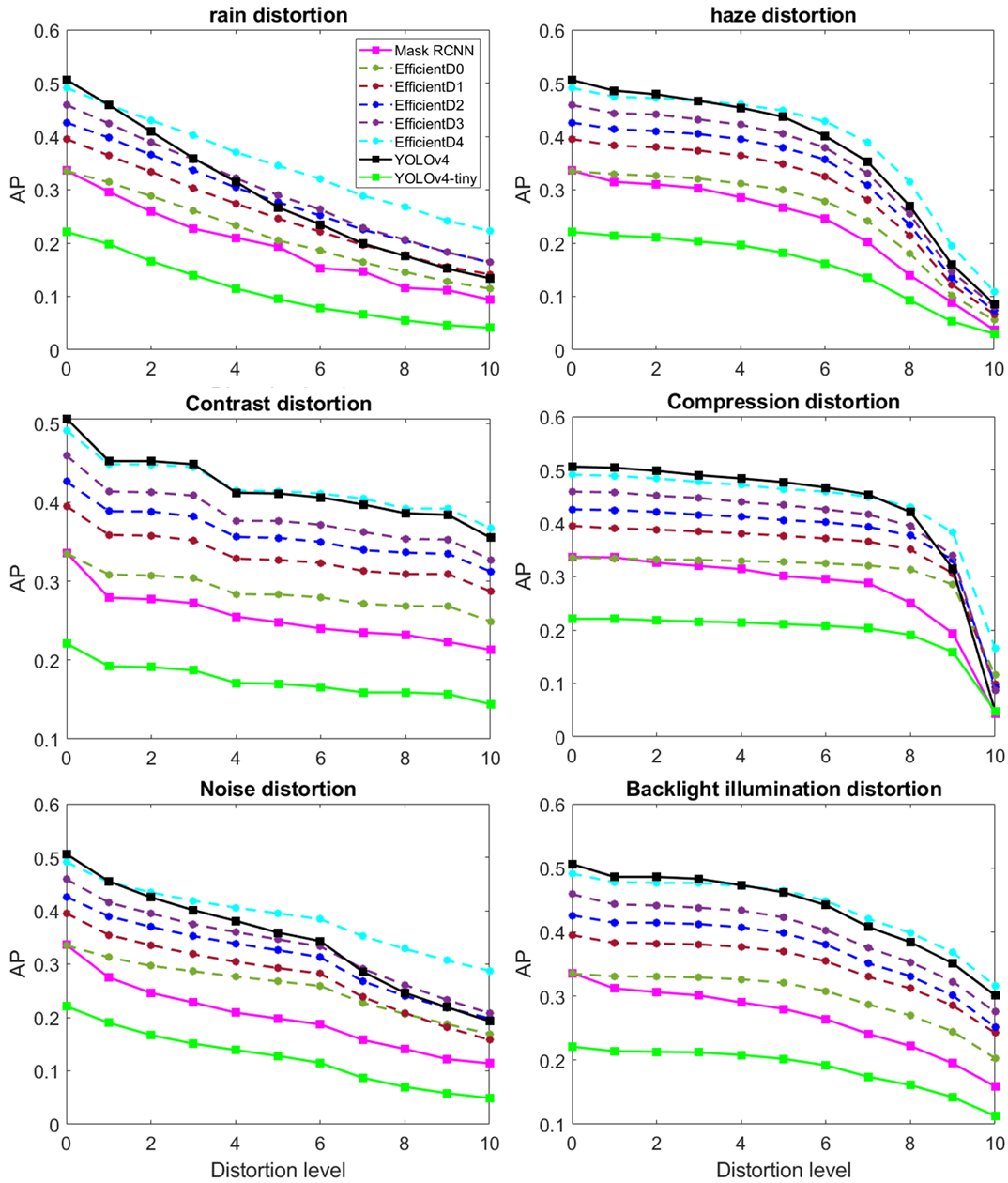}
    \caption{Robustness of object detection models against distortions.}
    \label{eval1}
\end{figure}
Figures \ref{eval2} and \ref{eval1} illustrate the evolution of the AP score of the considered OD models according to each distortion type and level. These results show that the distortions reduce object detection performance of the OD methods by 20\% to 89.2\%. Moreover, they highlight the weaknesses of YOLOv4 method against distortions (strongly decreasing curves in figures \ref{eval1} and \ref{eval2}), which needs a deeper study. 
The evaluations in figure \ref{eval2} indicates that the global distortions have more impact than local distortions on OD performance. While local and atmospheric disturbances have a roughly uniform impact on OD methods despite their different architectures, the other distortions produce effects separate and independent. Indeed, overall models performance are greatly affected by distorted images but not uniformly. Each type of distortion has a different rate of impact on models performance depending on the distortion level and OD method. 
In order to better evaluate the impact of distortions on OD models, we computed the robustness rate that represents the ratio between mAP scores of the models related to each distortion type and level, and non-distorted images (level 0).
This robustness rate is illustrated in (see figure \ref{rate})  using violin plot where the most relevant statistical features are highlighted.
Results from figures \ref{eval2}, \ref{eval1} and \ref{rate} indicate that the EfficientDet method's robustness decreases despite its network depth (D0 to D3).
Furthermore, the YOLOv4-tiny model was the least robust to distortions with an average decrease with  AP score of 36.7\% between the distortion levels 0 and 10 (lowest average robustness rate at 63.3\%, see mAP score in figure \ref{rate}). 
This observation lead us to choose this model to evaluate the improvement in robustness using a data augmentation as proposed in our dataset.

\subsection{Training on distorted dataset}
This section presents experimental results on MS-COCO 2017 evaluation set with 5K images. YOLOv4-tiny model is trained following the protocol in MS-COCO challenge, i.e., using the trainval35k set for training that includes 80k images from the train set and 35k images from the validation set. 
\begin{table}[h]
\caption{Object detection percentage performance of the YOLOv4-tiny trained model on our distorted dataset.}
\begin{center}
\begin{tabular}{p{1.6cm}|p{0.7cm}p{0.7cm}p{0.6cm}p{0.6cm}p{0.6cm}||p{0.8cm}}
\hline
Distortion&\multicolumn{5}{c||}{Distortion level (\%)}&mAP\\
 Type&2&4&6&8&10& Average\\
\hline\noalign{\smallskip}
\hline
Noise&9.58&22.3&37.9&80.3&114&47.2\%\\
Contrast&-0.54&1.8&2.47&3.23&5.71&1.99\%\\
Compression&-2.36&-0.48&-0.5&4.84&35.4&4.26\%\\
Rain&12.7&40.4&71.4&101.8&114.3&61.7\%\\
Haze&-0.49&3.14&10.5&29.8&51.7&15.8\%\\
Motion-Blur &6.11&39.5&95.3&151.4&183.3&86.3\%\\
Defoc-Blur&1.02&14.2&22.4&26.3&26.7&16.5\%\\
Loc. MBlur&10.3&31.6&46.9&59.3&67.0&39.8\%\\
Loc. Defoc.&3.08&16.0&23.3&25.5&25.9&17.2\%\\
BackLight&1.45&5.85&19.1&40.3&76.8&24.1\%\\
\hline
\hline
Average&\multirow{2}{*}{4.11\%}&\multirow{2}{*}{17.9\%}&\multirow{2}{*}{32.9\%}&\multirow{2}{*}{52.3\%}&\multirow{2}{*}{70.1\%}&\multirow{2}{*}{\textbf{31.5\%}}\\
 per level&&&&&&\\
\hline
\end{tabular}
\end{center}
\label{Training}
\end{table}
We first performed this training with the original MS-COCO trainval35k set to obtain reference values, then training with the distorted dataset, which allowed us to evaluate the contribution of the data augmentation on the robustness of the model. Each distortion type represents 5\% of the database used for the training process, i.e., 5.9K images for each distortion.
In our case, the hyper-parameters are taken from the YOLOv4 method but with some tuning.
The training step is 500,050, with the initial learning rate multiplied by 0.1 at 400,040 step then 450,045 step. 
The evaluation of the training improvement is achieved by comparing the evaluation scores of both models trained on original and distorted train sets, respectively, on the distorted MS-COCO validation sets for each distortion level. These results shown in  table \ref{Training} (Columns 2-6) are obtained by computing the ratio between AP COCO scores of the distorted model on the natural model.
We observed that the training has a significant impact on the robustness improvement with an average increase of 31.5\% of performance (see last column in table \ref{Training}).
Furthermore, the study highlighted that the training improves robustness for high-level distortions (see last row in table \ref{Training}).
\begin{table}[h]
\caption{Object detection performance of the YOLOv4-tiny trained model on our distorted natural sub-sets.}
\begin{center}
\begin{tabular}{p{1.9cm}|p{1.15cm}p{1.15cm}p{1.15cm}p{1.15cm}}
\hline
Natural&\multirow{2}{*}{AP}&\multirow{2}{*}{rel. AP}&\multirow{2}{*}{mIoU}&\multirow{2}{*}{rel.mIoU}\\
 distortions&&&&\\
\hline\noalign{\smallskip}
\hline
Noise&0.003&0.99\% &0.005&0.65\%\\
Contrast&0.011&5.47\%&0.007&0.95\%\\
Motion-Blur&-0.012&-5.17\% &-0.004&-0.54\%\\
Defocus&0.015&12.9\%&0.0&0.0\%\\
Rain&-0.009&-2.81\% &0.006&0.79\%\\
Loc. MBlur&-0.006&-3.13\%&0.0&0.0\%\\
BackLight&0.011&8.94\% &0.005&0.68\%\\
\hline
\end{tabular}
\end{center}
rel.: ratio between scores from distorted and original images, mIoU: mean Intersection over Union.
\label{Training2}
\end{table}
It can be observed from the results of  Table \ref{Training2}, that training has a weak impact on robustness against natural distortions with an average improvement of 2.5\% for the mAP score. 
However, the interpretation of these results deserves further investigations, taking into account the fact that the distortions present in the original database are of low amplitude \cite{Begh2022DN-COCO}. Indeed, the proposed data augmentation makes use of  images with higher amplitude distortion (see average per level of column 2 in table \ref{Training}).
 
\section{Conclusions and Future Work}
This study revealed the impact of the quality of the acquired images in the development of OD methods. Indeed, through this study we have shown that increasing the training database with complex scenarios containing different distortions improves the performance of OD models. 
We have also shown that this is true even for the best state-of- the-art methods. 
We, particularly,  propose an original strategy of image distortion generation applied to the MS-COCO dataset that combines some local and global distortions to reach a much better performance of OD process. We have shown that training with this distorted dataset improves the robustness of models by up to an average of 31,5\%. We, also,  provide a custom dataset including the natural images distorted from MS-COCO dataset to perform a more relevant evaluation of the robustness concerning distortions. 
The database and the generation source codes of the different distortions is made publicly available on github plateform.
Considering dataset with local and global distortions is an interesting way for future research.
\label{conclusion}



\bibliographystyle{IEEEbib}
\bibliography{refs}

\end{document}